\documentclass[fleqn,10pt]{wlscirep}
\usepackage[utf8]{inputenc}
\usepackage[T1]{fontenc}

\usepackage{booktabs}
\usepackage{amsmath}
\usepackage{algorithm}
\usepackage{algpseudocode}
\usepackage{booktabs}

\usepackage{xcolor}
\usepackage{colortbl} 
\usepackage{multirow}
\usepackage{makecell}

\title{Detection of ChatGPT Fake Science with the xFakeSci Learning Algorithm}

\author[1,2,+,*]{Ahmed Abdeen Hamed}
\author[3,4,+,*]{Xindong Wu}

\affil[1]{Complex Adaptive Systems and Computational Intelligence Laboratory, State University of New York at Binghamton, Binghamton, New York 13902, United States}
\affil[2]{Sano Centre for Computational Medicine, Clinical Data Science, Cracow, 30-072, Poland}
\affil[3]{Hefei University of Technology, Key Laboratory of Knowledge Engineering with Big Data (the Ministry of Education of China), Hefei, 230009, China}
\affil[4]{China Electronics Data Industry Group, Shenzhen, 518057, China}

\affil[*]{a.hamed@sanoscience.org, xwu@hfut.edu.cn}

\affil[+]{these authors contributed equally to this work}


\begin{abstract}
Generative AI tools exemplified by ChatGPT are becoming a new reality. This study is motivated by the premise that ``AI generated content may exhibit a distinctive behavior that can be separated from scientific articles''. In this study, we show how articles can be generated using means of prompt engineering for various diseases and conditions. We then show how we tested this premise in two phases and prove its validity. Subsequently, we introduce xFakeSci, a novel learning algorithm, that is capable of distinguishing ChatGPT-generated articles from publications produced by scientists. The algorithm is trained using network models driven from both sources. To mitigate overfitting issues, we incorporated a calibration step that is built upon data-driven heuristics, including proximity and ratios. Specifically, from a total of a 3952 fake articles for three different medical conditions, the algorithm was trained using only 100 articles, but calibrated using folds of 100 articles. As for the classification step, it was performed using 300 articles per condition. The actual label steps took place against an equal mix of 50 generated articles and 50 authentic PubMed abstracts. The testing also spanned publication periods from 2010 to 2024 and encompassed research on three distinct diseases: cancer, depression, and Alzheimer's. Further, we evaluated the accuracy of the xFakeSci algorithm against some of the classical data mining algorithms (e.g., Support Vector Machines, Regression, and Naive Bayes). The xFakeSci algorithm achieved F1 scores ranging from 80\% to 94\%, outperforming common data mining algorithms, which scored F1 values between 38\% and 52\%. We attribute the noticeable difference to the introduction of calibration and a proximity distance heuristic, which underscores this promising performance. Indeed, the prediction of fake science generated by ChatGPT presents a considerable challenge. Nonetheless, the introduction of the xFakeSci algorithm is a significant step on the way to combating fake science. 
\end{abstract}

\begin{document}
\flushbottom
\maketitle
%
%
\thispagestyle{empty}


\section{Highlights}
\begin{itemize}

    \item Proposing a prompt-engineering algorithm aimed at generating simulated biomedical abstracts pertinent to three distinct diseases, namely Alzheimer's, cancer, and depression.
    
    \item Presenting xFakeSci, a network-driven learning algorithm tailored to discern the authenticity of biomedical articles. The algorithm is trained using textual data, and calibrated using data-driven heuristics encompassing factors such as ratios and proximity metrics.
    
    \item Evaluating and benchmarking the algorithm using some of the most common classical  algorithms (e.g., NaiveBayes and Support Vector Machines) as a baseline. Strikingly, the xFakeSci algorithm demonstrates a notable disparity in performance compared to existing classical approaches.

\end{itemize}

\section{Introduction}
With Large Language Models (LLMs) and generative AI tools (e.g., ChatGPT) \cite{chatgpt} becoming a new reality, our world finds itself in a state of controversy. On one hand, there exists a camp of optimists who perceive their potential and seek to harness them. On the other hand, there are doubters who remain skeptical, seeking validation and further assessments to discern how this new paradigm will impact our lives. This division provides strong motivation for this study and catalyzes efforts towards providing a tool that assesses the capability of generating fake science by ChatGPT. Undoubtedly, real science, documented in scientific publications, stands as one of the most sacred sources of knowledge due to its invaluable contribution to future discoveries \cite{synnestvedt2005citespace,holzinger2013graph,usai2018knowledge}. Historically, the widespread of various predatory journals has fueled the rise of fake science \cite{thaler2015fish,hopf2019fake,ho2022let,frederickson2022addressing}, with the influence of social media exacerbating its impact \cite{rocha2021impact,walter2021evaluating}. Particularly during the global Coronavirus pandemic, the dissemination of misinformation regarding the importance of vaccination has led to its rejection by certain groups, endangering public health \cite{loomba2021measuring,lewandowsky2012misinformation,myers2009misinformation}. Another alarming incident during the pandemic involved the spread of an article reporting fake findings linking vitamin D deficiency to the death of 99 percent of the studied population. Although retracted, the misinformation was amplified by a DailyMail news article, spreading globally \cite{dailymail,abdeen2021fighting}. It is imperative to safeguard the authenticity of scientific publications from fraud or any influential factors that compromise the integrity of this crucial source of knowledge \cite{hamed2024safeguarding}.

In this article, we demonstrate how the emergence of ChatGPT (alongside many other generative AI tools) has impacted our society today: (1) the launch of many special issues and themes to study, assess, analyze, and test the impact and potential of ChatGPT \cite{eysenbach2023role,ieee_si_ed,finc_innov,mdpi_si_langs,oxford_sfi}, (2) the adoption of new policies by journals regarding ChatGPT authorship \cite{jmir_glm,pnas_2023,brainard2023scientists,fuster2023jacc,flanagin2023nonhuman,hamed2024safeguarding}, (3) the development of ChatGPT plugins and inclusion in professional services such as Expedia and Slack \cite{chatgpt_plugins}, and (4) the creation of educational tools (e.g., Wolfram) and the potential development of learning and educational support tools (e.g., Medical Licensing Examination \cite{gilson2023does}).

\subsection{Literature Background}
Due to the significance of fake science and the imperative for plagiarism detection mechanisms, numerous researchers have investigated fake science detection and employed diverse methodologies. Here, we list some of the most relevant works.

Chaka addressed the detection of content generated by tools such as ChatGPT and assessed it in five different languages. The work used a prompt-engineering approach to generate content in languages including Spanish, French, and German. The research concluded that detecting machine-generated content poses a significant challenge, and further investigations are much needed to protect against plagiarism \cite{chaka2023detecting}.

Cingillioglu also tackled the problem of identifying AI-generated essays using the Support Vector Machine algorithm (SVM) \cite{vapnik1999overview}. The work reported a 100\% accuracy for human-created articles; however, their approach failed to state the accuracy of ChatGPT-generated essays \cite{cing2023detecting}.

Elkhatat et al. evaluated 15 paragraphs generated by ChatGPT (versions 3.5 and 4.0) using means of other automatic detection tools such as Copyleaks \cite{copyleaks} and CrossPlag \cite{crossplag}. The authors concluded that AI detection tools were able to predict ChatGPT-generated content in isolation without mixing with other content. However, when the generated content was perturbed using human-written responses, all detection tools failed to produce accurate or consistent results \cite{elkhatat2023evaluating}. 

In their editorial effort, Anderson et al. also presented similar issues related to ChatGPT-generated articles. They further noted the lack of potent algorithms to detect AI-generated content. The authors concluded by mandating the need for publishers to introduce detection tools throughout the lifecycle of publications \cite{Andersone001568}.

Rashidi et al. presented their tool that determines whether an article is human-created or otherwise. Similar to ours, the study considered publication abstracts, which were collected from top-quality journals in the period of 1980-2023. Using a text-generated AI detector, the tool identified 8\% of real publications as machine-generated. The authors concluded their work by affirming the significance of advancing such research directions in the effort of protecting good science \cite{rashidi2023chatgpt}.

With this study, we respond to the urgent call to combat fake science by presenting the xFakeSci algorithm. The algorithm is primarily designed to detect ChatGPT-generated content and distinguish it from real PubMed abstracts of published articles. The algorithm operates in two modes: (a) a Single mode, where it is trained from a single source to predict one class, and (b) a Multi mode where the algorithm processes various types of resources to predict the correct label for each class. As a network-driven algorithm, it is trained by a model that utilizes the largest connected components (LCC) as an admissible heuristic. To mitigate overfitting issues, we implemented a data-driven calibration step that ensures the accurate prediction of documents, utilizing a small sample of training articles (100 articles from each source). In the Methods section, we will provide detailed explanations of the following components: (1) the prompt-engineering process of how fake documents were generated in three different diseases using ChatGPT, (2) the phases of testing the premise that content generated from ChatGPT may exhibit certain characteristics that reveal its identity and make it distinguishable from real science, (3) the xFakeSci algorithm, which we designed to predict the class of a given document and determine whether it is real or fake. We outline the computational steps, including constructing the network training models, calibrating using data-driven heuristics, testing the algorithm in multiple contexts (diseases) and data from various publication periods, and finally, (4) benchmarking it against the most common classical data mining algorithms as a verification step.

\section{Results}
 \subsection{Outcome of Evaluating the Premise}
As mentioned earlier, we investigated the premise that of whether AI-generated content may exhibit unique characteristics that differ from those observed in scientific articles. We tested this intuition two stages:

\subsubsection{Phase I: Analysis of Topological Properties of Network Training Models}
We constructed two types of network training models: one derived from content generated through prompt-engineering with ChatGPT, and the other from PubMed abstracts. We examined the structural properties of these network models in terms of the number of nodes and edges. These analyses were conducted within the contexts of three diseases: Alzheimer's, cancer, and depression. The node counts computed from ChatGPT training models were 519, 559, and 577, respectively. In contrast, the number of nodes generated from scientific publications varied across different time periods: for the years 2010-2014, it was 742, 755, and 801; for 2015-2019, it was 774, 828, and 755; and for 2020-2024, it was 817, 817, and 790. Regarding edge counts, ChatGPT training models exhibited 1194, 1050, and 1108 edges, whereas publication network models produced 861, 803, and 878 edges for the years 2010-2014; 940, 977, and 826 for 2015-2019; and 958, 1030, and 809 for 2020-2024.

\begin{figure}[ht]
    \centering
    \includegraphics[width=0.75\textwidth]{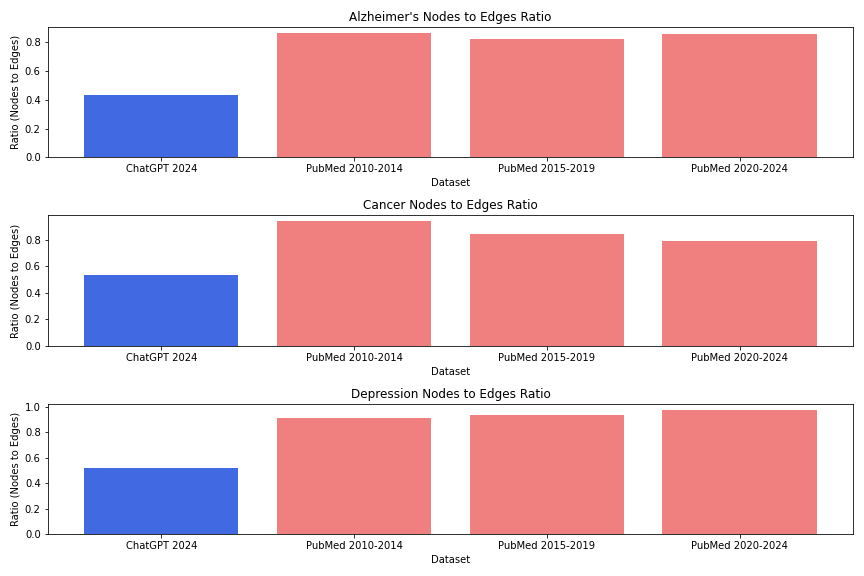} 
    \caption{Nodes to Edges Ratios for Different Datasets ChatGPT vs Scientific Articles}
    \label{fig:premise1}
\end{figure}

These findings, presented in Table \ref{tab:model}, suggest that ChatGPT-generated datasets generally have fewer nodes compared to scientific articles. However, our analysis also revealed that ChatGPT network models tend to have a higher number of edges relative to publication datasets. This observation is visually depicted in Figure \ref{fig:premise1}, which highlights the strikingly lower node-to-edge ratios of ChatGPT models compared to network models derived from scientific articles.

\subsubsection{Phase II: Further Testing the Distinctive Behavior in ChatGPT-Generated Documents}
\begin{figure}[ht]
  \centering
  \begin{minipage}{.5\textwidth}
    \centering
    \includegraphics[width=\linewidth]{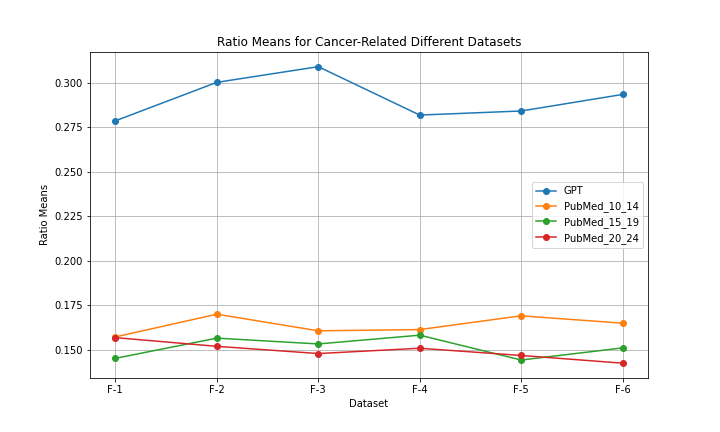}
    \caption{Comparison of Calibrating Ratio Means for the Cancer Disease.}
    \label{premise2a}
  \end{minipage}%
  \begin{minipage}{.5\textwidth}
    \centering
    \includegraphics[width=\linewidth]{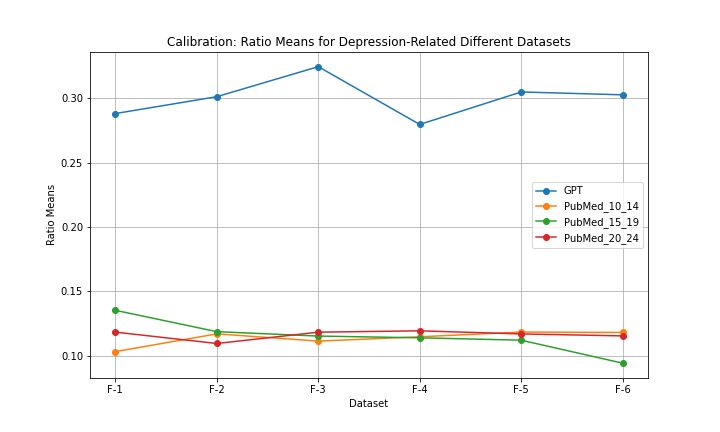}
    \caption{Comparison of Calibrating Ratio Means for the Depression Disease.}
    \label{premise2b}
  \end{minipage}
\end{figure}

To further investigate the premise, we conducted a test to analyze the mean ratios of contributing bigrams extracted from k-Folds against the document word count. This analysis aimed to establish a baseline for assessing the contribution of bigrams to the overall content structure. The results revealed a consistent pattern across all three disease datasets. Specifically, ChatGPT-generated datasets exhibited significantly higher ratios than their scientific publication counterparts in each of the k-Folds used. For instance, in the Alzheimer's disease dataset, ChatGPT scores were (0.27, 0.30, 0.30, 0.28, 0.28, 0.29), while scientific publications from 2010-2014 scored (0.16, 0.17, 0.16, 0.16, 0.17, 0.16), for 2015-2019 (0.15, 0.16, 0.15, 0.16, 0.14, 0.15), and for 2020-2024 (0.15, 0.15, 0.14, 0.15, 0.14, 0.14). These findings are consistent across the other two diseases, as evident in Table \ref{tab:ratio_means}. Figures \ref{premise2a} and \ref{premise2b} clearly demonstrate that the k-Folds ratios calculated from ChatGPT-generated data are significantly higher than those derived from scientific publications across different years and scopes. They further illustrate a similar pattern for the cancer and depression datasets. This evidence reinforces the notion that ChatGPT-generated content may exhibit distinct characteristics compared to scientific articles.

\subsection{Outcome of Label Prediction of Multi-Mode Classification Experiments}

To establish confidence in our method and ensure consistent performance of the xFakeSci algorithm, we conducted two types of experiments in the subject area of three different diseases. Additionally, we performed experiments to evaluate whether the year of publication plays a role in class prediction. This section presents the outcomes of experiments utilizing ChatGPT-generated text obtained algorithmically using ChatGPT prompt-engineering, as outlined in Algorithm \ref{algo:prompt-eng}, and scientific publications retrieved from the PubMed web portal \cite{pubmed-portal} related to the Alzheimer's, cancer, and depression diseases.

Here, we present the results of multi-mode experiments, where xFakeSci was trained using a combination of ChatGPT and PubMed abstracts and evaluated on a dataset of unseen documents from all three diseases. Specifically, we trained xFakeSci using an equal-sized dataset of ChatGPT-generated and PubMed abstracts. Then, we calibrated the algorithm using the exact number of k-Folds for each disease. For the PubMed dataset, we used abstracts of articles published between 2020 and 2024.

\begin{figure}[h]
    \centering
    \includegraphics[width=.95\linewidth]{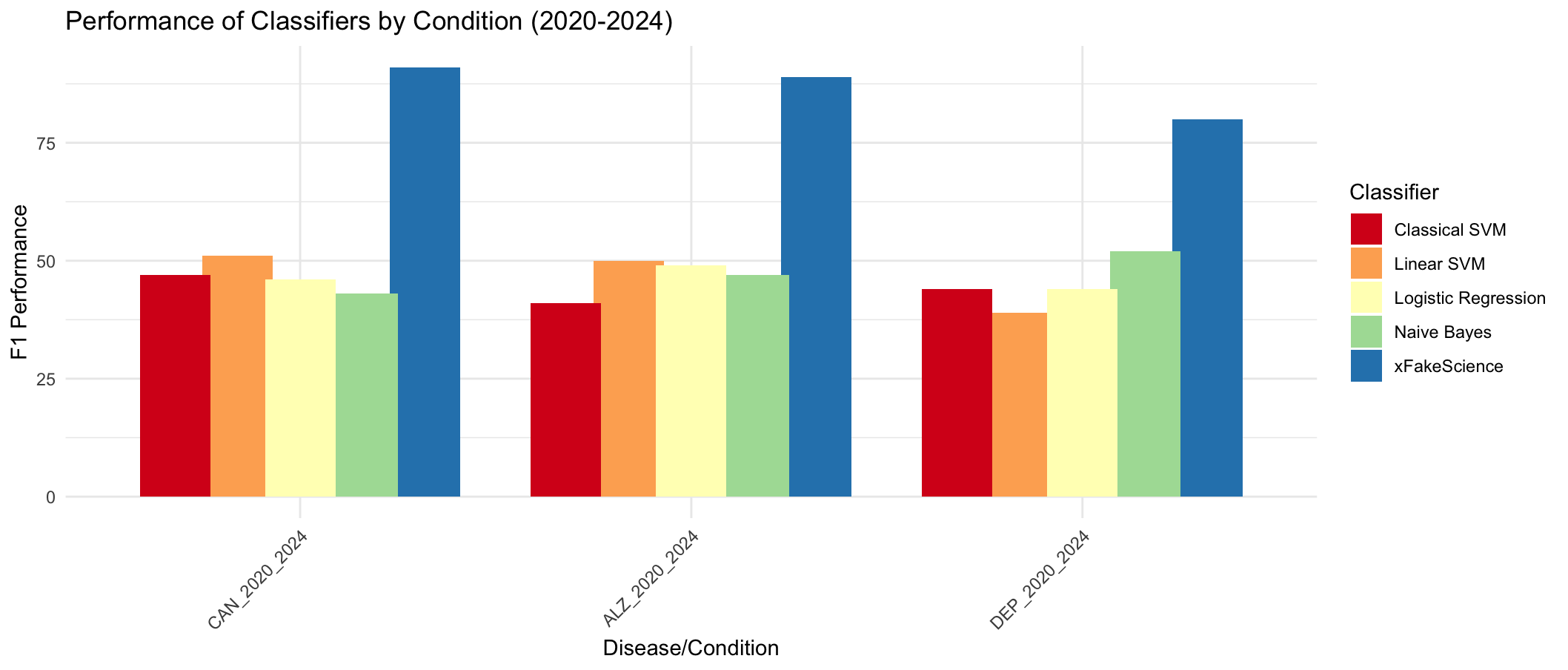}
    \caption{Multi-Mode Experiments: F1 Classification Scores for (Cancer, Alzheimer's, and Depression) for publications gathered in period (2020-2024.)}
    \label{fig:f1-diseases}
\end{figure}

For each disease, we tested xFakeSci on 100 articles, comprising 50 PubMed abstracts and 50 ChatGPT-generated documents. Table \ref{tab:f1-by-disease} summarizes the performance of xFakeSci in this mode, capturing the number of true positives (TP), true negatives (TN), false positives (FP), and false negatives (FN) from both the publication and ChatGPT test items. Using the F1 measure, we note that xFakeSci scored 80\%, 91\%, and 89\% for depression, cancer, and Alzheimer's, respectively. Figure \ref{fig:f1-diseases} provides a comprehensive analysis of the classification results.

Table 6 demonstrates that xFakeSci detected all 50 PubMed publications for each disease (TP=50). Additionally, we observed that the algorithm identified the ChatGPT-generated documents to varying extents (TN=25, 41, 38) for depression, cancer, and Alzheimer's, respectively. It remains concerning that ChatGPT is classified as PubMed with (FP=25), indicating that 50\% of the test documents are misclassified as real publications. Further research is needed to investigate and improve performance.

\subsection{Outcome of Publication Period as a Factor for the Multi-Mode Classification Experiments}
The purpose of this section is to test whether the publication period is a factor in making predictions and assigning labels to a dataset of documents with mixed classes (ChatGPT vs PubMed articles). Here, we use the F1 metric as a measure to present our results and explain the associated performance of our algorithm (captured in Table \ref{tab:f1_scores}). For each disease dataset extracted from three periods (2020-2024; 2015-2019; 2010-2014), we computed the F1 score.

For the cancer disease dataset, the F1 scores recorded were 91\%, 92\%, and 94\% for the three different periods. These scores show a consistent improvement in predicting the labels with older publication datasets, from 2010 to the present.

For the depression disease dataset, the F1 score remained constant over time at 80\%, indicating that the pattern did not show deterioration over time.

For the Alzheimer's disease dataset, the scores showed a slight improvement from 89\% (in 2020-2024) to 90\% in (2015-2019), but it dropped back to 89\% for the (2010-2014) disease dataset. While the pattern of prediction improvements did not hold as we analyzed older publications, the score did not degrade below the F1 score of 89\% for (2020-2024).

\subsection{Outcome of xFakeSci Performance Analysis against other Data Mining  Algorithms}
We compared the performance of xFakeSci against some of the most common state-of-the-art algorithms. Specifically, we conducted various performance evaluation experiments against the following algorithms: (1) Naive Bayes, (2) Support Vector Machine (SVM), (3) Linear Support Vector Machine (SVM), and (4) Logistic Regression. Some of these algorithms are listed among the Top-10 Data Mining Algorithms \cite{wu2008top}. To establish fairness, we trained each of the algorithms using the exact training dataset used against xFakeSci in multi-mode (where we train and test with mixed datasets). For training, we used the first 100 PubMed abstracts and the first 100 documents of the ChatGPT-generated dataset. As for testing, we used a combination of 50 PubMed abstracts followed by 50 ChatGPT-generated documents. 

\begin{figure}[h]
    \centering
    \includegraphics[width=0.95\linewidth]{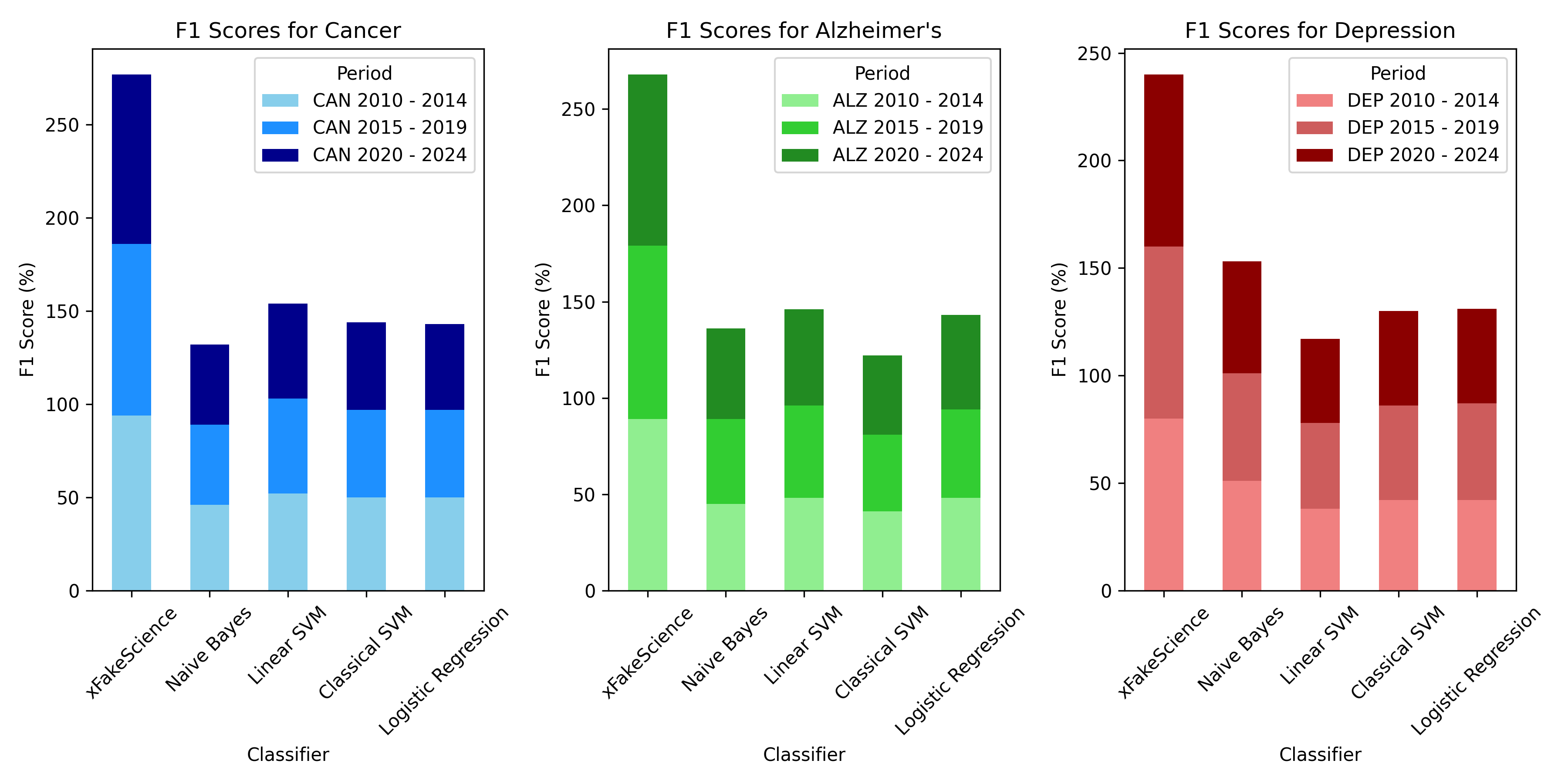}
    \caption{Accuracy of xFakeSci vs Classical Data Mining Algorithms Showing Publication Year as a Prediction Factor.}
    \label{fig:year-factor}
\end{figure}

Each algorithm was used as a blackbox and received input from two sources (training and test), and in turn, produced a detailed analysis in the form of (TP, TN, FP, FN). Using these metrics, the F1 score was computed accordingly. Figure \ref{fig:year-factor} visually depicts the F1 scores observed over time for the 5 algorithms (including xFakeSci) in three different diseases, presented by sub-figures, one for each disease. Each sub-figure shows stacked bars for the three periods of publications, and each bar represents the F1 score resulting from a given algorithm. Table \ref{tab:f1_scores-most-recent} captures the performance analysis of the xFakeSci algorithm against other classical data mining algorithms (Naive Bayes, Linear SVM, Classical SVM, and Logistic Regression). The performance is presented using the F1 metric for publications spanning between 2020 and 2024. The table shows that xFakeSci scores ranged from 80\% to 91\%, while those of the data mining algorithms fluctuated between 43\% and 52\%.

\begin{figure}[h]
    \centering
    \includegraphics[width=0.95\linewidth]{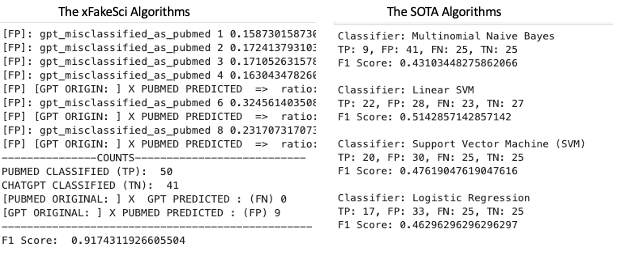}
    \caption{Showing (TP, TN, FP, FN) Mertics and F1 Scores for xFakeSci vs Classical Data Mining algorithms Using Cancer Dataset From 2020-2024}
    \label{fig:sota-evals}
\end{figure}

Moving to the period of 2014 to 2019, xFakeSci demonstrated F1 scores ranging from 80\% to 92\%, compared to other data mining algorithms which exhibited F1 scores in the range of 43\% to 51\%. Lastly, during the period of 2010 to 2014, the F1 scores achieved by xFakeSci fluctuated between 80\% and 94\%, whereas the F1 scores of the other data mining algorithms were recorded between 38\% and 52\% as shown in Table \ref{tab:f1_scores}. Figure \ref{fig:sota-evals} shows a screenshot providing evidence of achieving 91\% accuracy measured by the F1 metric, while the F1 scores of the other data mining algorithms fluctuated from 43\% to 51\%. All three sub-figures show a consistent pattern where xFakeSci clearly outperforms all the other four algorithms. The F1 score is calculated using Equation \ref{eq:f1}, and the analysis was done using the scikit-learn library \cite{scikit-learn}.
\begin{equation}
\label{eq:f1}
    F1 = \frac{2 * \text{TP}}{2 * \text{TP} + \text{FP} + \text{FN}}
\end{equation}

\section{Methods}
\subsection{Data Collection}
We compiled two distinct types of datasets for the study: (1) \textbf{Literature dataset:} To establish a baseline for comparison and train the xFakeSci algorithm, we utilized the PubMed archive to retrieve scientific articles. We employed three search queries: (a) ``Alzheimer's disease and co-morbidities,'' which we generated 1196 JSON records, (b) ``cancer and co-morbidities'', generated 1243 JSON records, and (c) ``depression'' which generated 1513 JSON records. To assess the influence of publication year on fake science detection, we conducted these searches at five-year intervals, resulting in three distinct datasets for each disease (2010-2024). (2) \textbf{ChatGPT-generated dataset:} We obtained this dataset by programmatically prompting the ChatGPT API (version: 3.5, model: ``gpt-3.5-turbo-16k'') to generate simulated articles. The prompt-engineering process comprises two primary components:
\begin{itemize}
    \item \textbf{Prompt Engineering ChatGPT for Simulated Article Generation:} We employed ChatGPT, a generative AI tool, for generating simulated articles. We implemented a prompt-engineering technique to guide the process of text generation in three distinct diseases: Alzheimer's, cancer, and depression.
	
    \item \textbf{Predicting Fake Science:} We devised a network-centric learning algorithm, called xFakeSci, which we trained on text documents of published scientific articles and ChatGPT-generated documents. The algorithmic steps involved in this process are explained in the subsequent sections.
\end{itemize}

\subsection{ChatGPT Prompt Engineering for Article Generation}
We utilized the ChatGPT API (version: 3.5, model: ``gpt-3.5-turbo-16k'') to engineer prompts for generating simulated articles in the subject areas of real published articles (Alzheimer's, cancer, and depression). These prompts were parameterized using information from the real articles (search keywords used for article retrieval and the average number of words) to make them comparable to real abstracts. They included three key elements: (a) the role of the prompt: ``a biomedical researcher,'' (b) the request example: ``Generate a list of 20 simulated PubMed-style abstracts,'' (c) topic example: ``the Alzheimer's disease,'' and (d) specifications: each article must contain ID, Title, and Abstract fields. We also instructed prompts to generate a valid JSON response with these specifications. The number of words helped offset any bias and made the fake articles comparable to the precise level of detail required (the 200-250 words range is a common requirement by many prominent biomedical informatics journals). Table \ref{tab:result_counts} captures the search queries and the number of fake articles generated in the JSON format. 
\begin{table}[htbp]
\centering
\caption{Result Counts of Prompt Engineering}
\label{tab:result_counts}
\begin{tabular}{@{}lc@{}}
\toprule
Search Query & Count (JSON Records) \\ \midrule
Alzheimer's disease and co-morbidities & 1196 \\
Cancer and co-morbidities & 1243 \\
Depression & 1513 \\ 
\midrule
Total number of generated articles & 3952 \\
\bottomrule
\end{tabular}
\end{table}

The prompt-engineering process is computationally described in Algorithm \ref{algo:prompt-eng}. Although this process was done programmatically, due to the timeout limit, we executed it to produce 20 simulated articles at a time. This prompt-engineering approach enabled us to generate a large corpus of simulated articles that closely resembled real scientific publications in terms of structure, content, and overall style. This dataset played a crucial role in training the xFakeSci algorithm, enabling it to accurately distinguish between real scientific articles and machine-generated ones. 
\begin{algorithm}[!htbp]
\caption{The Computational Process of ChatGPT Prompt Engineering for Article Generation}
\label{algo:prompt-eng}
\begin{algorithmic}
    \Require diseases = [``Alzheimer's'', ``Cancer'', ``Depression'']
    \Require $article\_number$ = 20.
    \Require $abstract\_length\_range$ = (200-250) of words in each article.
    \For{For each disease $d$ in diseases}
        \State [System Role: ] You are a biomedical researcher specialized in studying $d$ disease.
        \State [Request Topic:] disease.
        \State [Request Content:] Generate a list of $article\_number$ simulated PubMed-style abstract.
        \State [Request Specs:] Documents must contain: ``GPT-ID, Title, and Abstract''; and abstract must be between 200-250 words''.
        \State [Request Details:] Provide disease and co-mobridities detailed information.
        \State [Response Format:] A valid JSON format returned as an array of valid JSON records.           
    \EndFor
    \State \textbf{Return} an array list of ChatGPT-generated article in JSON format.
\end{algorithmic}
\end{algorithm}

\subsection{Prediction of Fake Articles using xFakeSci}
The xFakeSci algorithm is a network-driven label prediction algorithm designed to distinguish between real scientific articles and machine-generated ones. This entails that the algorithm has two main tasks: training the model and testing it to detect the label of entirely new documents that have never been seen before. In this section, we introduce the computational steps that describe the prediction process, starting with (1) the construction of network training models, (2) the calibration of the algorithm, and (3) the label prediction for each of the ChatGPT-generated articles.

\subsubsection{Model Derivation and Network Construction}
Since our training model is network driven from text, we used the Term Frequency-Inverse Document Frequency (TF-IDF) \cite{aizawa2003information,qaiser2018text,ramos2003using,trstenjak2014knn,wu2008interpreting,zhang2011comparative} to extract word features as building blocks of the training models. The TF-IDF algorithm can be configured to generate two consecutive words (known as bigrams) that may prove significant across an entire dataset \cite{tan2002use,hirst2007bigrams}. Equation \ref{eq:tfidf} shows the mathematical representation of the TF-IDF. where \(\text{tf}(t, d)\) is the frequency of bigram \(t\) in document \(d\),and \(\text{idf}(t, D)\) is the inverse document frequency of bigram \(t\) in the document set \(D\). 

The term frequency of a term \(t\) is calculated as the ratio between the number of occurrences of the term divided by the total number of terms in a document  \(d\) as show in Equation \ref{eq:term-freq}. The inverse document frequency (\(\text{idf}(t, D)\)) is calculated as in Equation \ref{eq:idf} where \(N\) is the total number of documents in the collection and \(\text{df}(t, D)\) is the number of documents containing the bigram \(t\).

\begin{equation}
    \label{eq:tfidf}
    \text{TF-IDF}(t, d, D) = \text{tf-idf}(t, d, D) = \text{tf}(t, d) \times \text{idf}(t, D)
\end{equation}

\begin{equation}
    \label{eq:term-freq}
    \text{TF}(t, d) = \frac{\text{Number of occurrences of term } t \text{ in document } d}{\text{Total number of terms in document } d}
\end{equation}

\begin{equation}
    \label{eq:idf}
    \text{IDF}(t, D) = \log\left(\frac{N}{1 + \text{df}(t, D)}\right)    
\end{equation}

To construct a training model, we extracted bigrams to form a network model as follows: the individual words of a bigram served as nodes, and edges represented the relationship between the bigrams. To illustrate the utility of bigrams in constructing the training model, let's consider a scenario concerning the ``depression disease'': bigrams such as ``mental health,'' ``health condition,'' and ``condition worsen'' form connections based on the common words they share, enabling a network that can be analyzed for various purposes. Using this mechanism, we constructed two distinct training models: one from the abstracts of published literature (labeled as the ``PUBMED'' class), and another from ChatGPT-generated text (labeled as the ``GPT'' class). To ensure fairness and prevent biases, both models were constructed from the same number of documents (100 abstracts and 100 ChatGPT-generated). Both datasets were processed using identical series of steps, including removing stopwords and sentence tokenization.

Algorithm \ref{alo:netcons} outlines the steps involved in building the network model from bigrams. We applied this algorithm twice, once to create a publication training model and another to create a ChatGPT model. We recorded the corresponding statistics (numbers of nodes and edges) for each model in Table \ref{tab:model}. The initial observations revealed a consistent pattern where models constructed from ChatGPT-generated text exhibited the lowest number of nodes, yet they also maintained the highest number of edges. The resulting models showed disconnected components and fragmented communities, requiring pruning. This need was satisfied by applying the Largest Connected Components (LCC) algorithm \cite{dorogovtsev2001giant}, which ensured that the resulting networks maintained high connectivity. The LCC presents an admissible pruning heuristic due to the presence of high-degree nodes that promote network stability and robustness \cite{kitsak2018stability, Beygelzimer2005, Zhang2017, Bellingeri2020}.
\begin{algorithm}[!htbp]
\small
\caption{Network Construction from Bigrams Computed from Text Documents (PubMed, and ChatGPT)}\label{alo:netcons}
\begin{algorithmic}[1]
\Require
  \Statex $\mathcal{D}$: [document dataset (PubMed or ChatGPT))]  
  \Statex $\mathcal{T}$: training graph
  \Statex $\mathcal{B}$: empty list of bigrams
\Ensure
  \Statex $\mathcal{G}$: fully populated graph
 \For{each document $d$ in $\mathcal{D}$ dataset (either from ChatGPT or PubMed)}
  \State $B$ $\leftarrow$ Compute Term Frequency - Inverse Term Frequency tf-idf($d$)
  \For{each bigram $\mathcal{b}$ in $\mathcal{B}$ encoded as source and target $b(s, t)$}
    \State Form an edge $e(s, t)$ for the unigram constituents of the bigram $b$.
    \If{$e$ does not exist in the training graph: $\mathcal{T}$}
      \State Add $e$ to the training model $\mathcal{T}$.
    \EndIf
    \State Reset $B$ $\leftarrow$ [] to populate with next document
  \EndFor
\EndFor
\State \textbf{return} $\mathcal{T}$  a fully populated graph from dataset bigrams
\end{algorithmic}
\end{algorithm}

\subsection{Evaluating the Premise of ChatGPT's Distinctive Behavior}

As mentioned earlier, the training models are constructed from the first 100 articles from each dataset. To test the premise of how ChatGPT may exhibit distinctive behavior, we divided the remaining articles into k-Folds, each containing 100 articles. The main idea of such a test was to measure the impact of each fold on the corresponding training model, specifically how the bigrams extracted from each of the folds altered the Largest Connected Components (LCCs) of their respective data types.

For the Alzheimer's disease dataset, we constructed three training models: the impact of bigrams was determined by calculating the mean ratio between the number of bigrams contributing to the LCC and the total number of words in each article within a fold. This process is captured by measuring the average contribution rate of the bigrams of a given fold. Algorithm \ref{algo:calib_ratios} provides the pseudocode for this step.
\begin{algorithm}[!htbp]
\caption{Compute the Mean of Ratios for a Corresponding Fold} \label{algo:calib_ratios}
\begin{algorithmic}[1]
\Require Training model $M$, 
\Require One fold of 100 articles from $k$-folds, 
\Require Computed ratios of each fold $R \leftarrow []$ \Comment{Initialize as an empty list}
\Ensure Fold mean
\For{each document $d$ in $D$}
    \State $\text{doc\_ratio} \leftarrow 0.0f$ \Comment{Initialize using a value of 0.0f}
    \State $\text{bigram\_count} \leftarrow 0$ \Comment{Initialize using a value of 0}
    \State $\text{doc\_wc} \leftarrow \text{compute\_word\_count}(d)$
    \State $B \leftarrow \text{Compute bigrams using TF-IDF}$
    \For{each bigram $b$ in $B$}
        \If{$b$ contributes an edge to the corresponding training model $M$}
            \State $\text{bigram\_count} \mathrel{+}= 1$ \Comment{Increment the count by 1}
        \EndIf
    \EndFor
    \State $\text{doc\_ratio} \leftarrow \frac{\text{bigram\_count}}{\text{doc\_wc}}$
    \State $R.\text{append}(\text{doc\_ratio})$ \Comment{Keep track of the ratios}
\EndFor
\State $\text{fold\_mean} \leftarrow \text{compute the mean of the ratios} \ \text{mean}(R)$
\State \Return{$\text{fold\_mean}$}
\end{algorithmic}
\end{algorithm}            

We summarized the analysis of each dataset and disease in Table \ref{tab:ratio_means}. The initial observations indicated that the ChatGPT ratios fluctuated between 27\% and 30\% for the Alzheimer's disease, 27\% and 29\% for cancer, and 28\% and 32\% for depression. In contrast, the ratios derived from scientific articles ranged between 14\% and 16\% for the Alzheimer's disease, 14\% and 17\% for cancer, and 9\% and 11\% for depression. The full analysis of these results will be discussed in the ``Results'' section. The analysis demonstrated that the ratios of ChatGPT-generated documents were significantly different from those computed from scientific articles. This distinction serves as proof that the premise is indeed true. Furthermore, this knowledge provides lower and upper bounds for each disease, offering more guidance to the algorithm to predict the label while avoiding issues of overfitting. Algorithm \ref{algo:calib} demonstrates how the ratios are computed. For brevity, we only demonstrate for the depression disease. Table \ref{tab:bound_ranges} presents the corresponding lower and upper bounds, which are necessary during the calibration phase.
\begin{algorithm}[!htbp]
\caption{Computing the Calibrating Ratio Ranges from k\_Folds}\label{algo:calib}
\begin{algorithmic}[1]
\Procedure{ComputeRanges}{$R, R'$}
    \State \textbf{Input:} List of ChatGPT k-Fold ratios $R$: $[r_0, r_1, r_2, ..., r_k]$
    \State \textbf{Input:} List of PubMed k-Fold ratios $R'$: $[r'_0, r'_1, r'_2, ..., r'_k]$
    \State \textbf{Output:} Dictionary of key/value pairs where the key is the name of the dataset, and the value is the lower/upper bound range.

    \State $gpt\_ranges \gets []$
    \For{each ratio in $R$}
        \State $gpt\_min \gets \min(R)$
        \State $gpt\_max \gets \max(R)$
        \State $gpt\_lower\_upper \gets \text{range}(gpt\_min, gpt\_max)$
        \State $gpt\_ranges.append(gpt\_lower\_upper)$
    \EndFor

    \State $pubmed\_ranges \gets []$
    \For{each ratio in $R'$}
        \State $pubmed\_min \gets \min(R')$
        \State $pubmed\_max \gets \max(R')$
        \State $pubmed\_lower\_upper \gets \text{range}(pubmed\_min, pubmed\_max)$
        \State $pubmed\_ranges.append(pubmed\_lower\_upper)$
    \EndFor

    \State \textbf{return} $[gpt\_ranges, pubmed\_ranges]$
\EndProcedure
\end{algorithmic}
\end{algorithm}

\subsubsection{Label Prediction of Articles: Real vs Fake}
Testing the premise in the above section demonstrated the fundamental differences in content behavior between fake ChatGPT articles and real publications. In this section, we present the learning algorithm, which is the main contribution of this paper. During the Coronavirus global pandemic, our previous work addressed the challenge of detecting fake news and science as an emerging infodemic \cite{abdeen2021fighting}. However, this work was limited by the lack of comprehensive machine-generated datasets that could adequately assess performance in the presence of fake data. Now, with the advent of ChatGPT and generative AI technologies, we can generate diverse datasets using prompt-engineering algorithms as demonstrated above. Additionally, the previous work did not make use of data-driven insights, which we incorporate as a calibration step. This is an intermediate phase that takes place after the training phase and before the label prediction phase. Due to these factors, the previous work was limited to a single-mode label prediction using a single type of dataset. Therefore, it was necessary to split the dataset into training and test sets. The following is a complete comparison of the previous work and the current features presented by xFakeSci, such as content type, configuration parameters, classification mode, calibration, and classification, as presented in Table \ref{tab:neonet_comparison}. 
\begin{table}[ht]
    \centering
    \caption{Comparison between NeoNet and xFakeSci Algorithms}
    \label{tab:neonet_comparison}
    \begin{tabular}{lcc}
        \toprule
        \textbf{Feature} & \textbf{NeoNet Algorithm} & \textbf{xFakeSci Algorithm} \\
        \midrule
        Content type & publication/news & publications/ChatGPT fake articles \\
        Calibration & not-supported & data-driven calibration \\
        Configuration Parameters & min-support and min-confidence & lower-bound/higher-bound ratio means \\
        Distance & no supported &  uses proximity distance for out-of-range instances\\        
        Classification & single-mode classification (news against news) & single/multi-mode classification (fake against real) \\
        Classification/Year & not-supported & comparative classification by the year of publication \\
        Multi-Mode Performance & not-supported & F1 score 80\%-94\% \\
        \bottomrule
    \end{tabular}
\end{table}

As shown in Table \ref{tab:neonet_comparison}, the xFakeSci algorithm is particularly designed to address multi-mode classification. Therefore, it is expected to train the algorithm using two or more independent types of data. Consequently, the algorithm also expects a hybrid test set of mixed types and will produce more accurate labels for each type. However, such modes suffer from what is known as the ``overfitting'' issue \cite{genkin2007large,feng2017overfitting,deng2019,khurana9691845_2023}. The introduction of the calibration step (by calculating the lower/upper bound ratios captured in Table \ref{tab:bound_ranges}) was to guide the decision of the final label prediction and avoid such an issue. The table demonstrates a clear separation of lower/upper bound ratios. Therefore, we further utilize such a mechanism by incorporating a calibrating step to further guide the classification process without having to train the algorithm with too many samples. The algorithmic steps for the calibration process are explained in Algorithm \ref{algo:calib}. Though the ranges provide an extra net for predicting the label, it is also possible that some document instances may fall outside the specified ranges of the datasets, which could result in not predicting a label correctly. Therefore, we introduced a proximity heuristic that favors the shortest distance to the ranges driven from the individual datasets (real or ChatGPT-generated) and assigns a label accordingly. Equations \ref{eq:distance} and \ref{eq:proximity} demonstrate how the distance is calculated.

\begin{equation}
    \label{eq:distance}
    \text{distance} = \min\left(\left| \text{point} - \text{range}_{end} \right|, \left| \text{point} - \text{range}_{start} \right|\right)
\end{equation}
\begin{equation}
    \label{eq:proximity}
    \text{proximity} = \min(\text{distance}_{d} , \text{distance}_{d'})
\end{equation}

Algorithm \ref{algo:xfakebibs} illustrates the computational steps for multi-mode execution, demonstrating the complexity involved, including the proximity distance. To use the algorithm in detecting fake science, it must be trained using two different types of data: (1) a real publication dataset and (2) ChatGPT-generated articles. The algorithm also expects the ratio means of each data source, which are computed using the calibration algorithm.
\begin{algorithm}[!htbp]
\caption{Multi-Mode Execution of the xFakeBib Algorithm}\label{algo:xfakebibs}
\begin{algorithmic}[1]
    \State \textbf{Input:} $[ChatGPT Model \ G, \  \ PubMed Model \ P]$
    \State \textbf{Input:} $[ChatGPT-ratios \ range_1, \ PubMed-ratios \ range_2]$
    \State \textbf{Input:} a dataset $D$ of mixed fake and real documents
    \For {each document $d$ in $D$ dataset}
        \State $ratio_1 \leftarrow \text{compute\_model\_contribution}(d, G)$
        \State $ratio_2 \leftarrow \text{compute\_model\_contribution}(d, P)$
        \If {$ratio_1$ in $range_1$}
            \State assign GPT as a label to document $d$
        \ElsIf {$ratio_2$ in $range_2$}
            \State assign PubMed as a label to document $d$
        \Else
            \State $dist_{range_1} \leftarrow \text{compute\_distance\_to\_range}(ratio_1, range_1)$
            \State $dist_{range_2} \leftarrow \text{compute\_distance\_to\_range}(ratio_2, range_2)$
            \If {$dist_{range_1} < dist_{range_2}$}
                \State assign GPT as a label to article $d$
            \ElsIf {$dist_{range_2} < dist_{range_1}$}
                \State assign PubMed as a label to article $d$
            \EndIf
        \EndIf
        \State capture stats $\leftarrow$ score prediction
    \EndFor
    \State \Return stats
\end{algorithmic}
\end{algorithm}

\section{Discussion}
In a world where generative AI has become widespread, various studies aimed to investigate the potential issues of using ChatGPT to generate fake science. The literature review showed a desperate need to advance the algorithmic approaches to discern real publications from fake ones, especially, when they are mixed. Our study aimed to address such issues incrementally. Specifically, we first tested the intuition of whether the content generated by ChatGPT may exhibit unique characteristics that distinguish it from real science. We explored this task using prompt engineering, where we created engineered datasets on the subjects of Alzheimer's, cancer, and depression diseases. In this work, we contributed a prompt-engineering algorithm on how to generate simulated content to evaluate this premise. Working with plain text (using publication abstracts or generated from ChatGPT), using the TF-IDF algorithm is a common approach to generate bigrams that can be used to construct more complex models. 

Our initial observation of networks generated from ChatGPT content is that they are highly connected and contain fewer nodes compared to networks constructed from real publication text. Additionally, when we calculated the ratios of the number of bigrams against the total number of words of documents on k-Folds, we found that the ratios of ChatGPT content are much higher than scientific abstracts. These two indications supported our intuition that ChatGPT documents exhibit distinguishable behavior than PubMed abstracts. One interpretation of this observation could be due to the inherent design of the ChatGPT engine. As observed, ChatGPT is optimized to generate highly convincing content by predicting the next correlated terms statistically using a Large Language Model. On the other hand, scientists prioritize accurate documentation of hypothese testing, scientific experiments, and careful explanation of observations. Describing science in terms of highly correlated words is not a goal of scientists. Clearly, the difference in goals may contribute to less connectivity in scientific publications.

Further, we introduced the xFakeSci algorithm, a learning algorithm that predicts a label for a given article. In the Methods section, we showed that it is designed to operate in two modes: (1) Single-mode: where only one type of articles from the same source is used for training and a new set of documents from the same pool is used for predicting the label of an article; and (2) Multi-mode: where the algorithm was trained from two sources and a hybrid train model (of real and generated datasets) was constructed to make the predictions. The single-model is trivial; therefore, we focused our experiments on demonstrating the multi-mode. We performed several experiments to do the following: (1) to test and measure how the xFakeSci algorithm predicts labels of ChatGPT generated documents for a given disease when mixed with scientific abstracts, (2) to evaluate whether the algorithm performs consistently using various datasets of different diseases not only one disease, (3) to test whether the year of the publication plays a role in predicting ChatGPT generated documents when mixed with publications from various periods (2020-2024, 2015-2019, 2010-2014), and (4) to benchmark the algorithm against a baseline of some of the most common data mining algorithms. Our results for each experiment used the TP, TN, FP, FN metrics and F1 scores. 

When testing whether the year of publication plays a role in label prediction, we observed F1 scores of 91\%, 92\%, and 94\% for cancer-related publications across different periods. This suggests a pattern of better detection of ChatGPT articles when mixed with older publications. However, identifying newer publications proved more challenging. For the Alzheimer's disease, while no improvement was observed, degradation was also absent. As mentioned earlier, the Alzheimer's datasets were the smallest among all datasets, limiting the calibration process due to fewer k-Folds compared to other diseases. In the case of depression, the algorithm exhibited consistent performance with an F1 score of 80\% across all periods. It's plausible that mental health data acquisition posed limitations, potentially constraining resources from this specific area. Testing this hypothesis involves measuring document similarity between PubMed and ChatGPT sources using lexical and semantic analysis.

Upon benchmarking xFakeSci against classical data mining algorithms, we observed an interesting pattern: xFakeSci correctly predicted all the scientific publications in all the experiments we performed. However, other algorithms misclassified publications as ChatGPT and vice versa (true positives, false positives, false negatives, and true negatives). xFakeSci, however, needed improvement in predicting true negatives (ChatGPT documents), as many ChatGPT documents were labeled as true positives (real publications). In all the experiments, the F1 scores of xFakeSci ranged between 80\% and 94\%. In contrast, the other data mining algorithms showed much lower performance, with F1 scores ranging between 32\% and 52\%. We attribute the high performance of xFakeSci to the calibration process, which was guided by ratios and proximity distances. Although the training model remained lightweight, both heuristics provided more guidance for predicting fake articles. This novel calibration method benefits from an abundance of data, without suffering from overfitting issues like other common classification algorithms. Clearly, the xFakeSci algorithm does not suffer from such a deficiency in identifying real articles when mixed with ChatGPT-generated content

While xFakeSci is designed to distinguish fake science from real, it can be applied to various types of text data, including clinical notes, clinical trial summaries, and interventions. With the widespread adoption of generative AI tools such as ChatGPT and Google Bard, ethical concerns may arise, such as clinicians using ChatGPT to generate clinical notes, potentially resulting in erroneous entries with serious consequences. In such cases, our algorithm may serve as a forensic tool to identify potentially fake portions of these reports.

While we have highlighted the potential for harm posed by ChatGPT and similar tools, it is also important to recognize their positive generative capabilities. For instance, ChatGPT played a crucial role in providing our algorithm with simulated data, which was essential for our work during the global pandemic in detecting fake news and publications \cite{abdeen2021fighting}. Moreover, ChatGPT can generate code snippets as building blocks for various basic tasks, including data visualization, across diverse programming languages. We are currently exploring this capability to construct workflows for life sciences applications. Additionally, the ChatGPT engine can effectively convert semi-structured content into popular formats like JSON, XML, and others. While these capabilities are undoubtedly useful, they necessitate the development of ethical standards to ensure responsible use of such tools.

Another intriguing potential use is that, when creatively engineered, ChatGPT could function as a valuable teaching assistant for academics and school teachers. It could potentially generate various ways to present questions while maintaining the integrity of the original content. Furthermore, ChatGPT could revolutionize scientific writing by providing support in addressing grammatical errors, typography, and paraphrasing, particularly for those whose native language is not English \cite{conroy2023chatgpt}.

\section{Conclusions and Future Directions}
When we asked a high school student about their knowledge of ChatGPT, they responded, ``Do you mean that tool that does my homework for you?'' Indeed, ChatGPT is an incredibly sophisticated tool with a wide range of impressive capabilities. Since the rise of ChatGPT, many new research topics have opened a new generative door, and many long-standing questions are now being investigated. However, the most significant concern associated with ChatGPT and other generative AI tools is that they could pose a threat to the future of science. If younger generations utilize ChatGPT to plagiarize, it could undermine the integrity of research and learning, potentially having a negative impact on the development of future pioneers.

While learning algorithms, such as xFakeSci, can assist in identifying fake science, there is an ethical obligation to use generative AI tools responsibly and regulate their usage \cite{hamed2024safeguarding}. It is worth noting that certain countries, such as Italy, have taken the extreme step of banning ChatGPT. While the authors believe such measures may be drastic, addressing ethical concerns is a new frontier that must be tackled. As ChatGPT itself states, ``It is up to individuals and organizations to use technology like mine in ways that promote positive outcomes and minimize any potential negative impacts.'' As advised by Anderson et al., it is also the responsibility of publishers and those involved in the production of science to play a proactive role in promoting good science. This includes raising awareness of the importance of implementing advanced fake science detection algorithms, including ours, and activating the use of technologies to distinguish fake research and fabricated findings \cite{Andersone001568}.

Looking ahead, there are several avenues for future research based on our current work: (1) conducting a preprocessing step (e.g., clustering) to group more closely related publications together (e.g., breast cancer, prostate cancer, and others), or separate diseases from co-morbidities. The use of knowledge graphs may be a powerful tool to use in continuing to investigate this research direction; (2) further experimentation in training and calibrating the xFakeSci algorithm by utilizing heuristics learned from preprocessing steps and the discoveries of clusters; and (3) testing the algorithm on more than two data sources (clinical reports, publications, and ChatGPT-generated documents).

\section*{Acknowledgments}
This research is supported by the European Union's Horizon 2020 research and innovation programme under grant agreement Sano No 857533 which is carried out within the International Research Agendas programme of the Foundation for Polish Science, co-financed by the European Union under the European Regional Development Fund, and created as part of the Ministry of Science and Higher Education’s initiative to support the activities of Excellence Centers established in Poland under the Horizon 2020 program based on the agreement No ``MEiN/2023/DIR/3796'', and the National Natural Science Foundation of China (NSFC) under grant 62120106008. The authors also acknowledge Laila Hamed for her valuable perspective on ChatGPT.

\section*{Author contributions statement}
A.H. conceived the idea(s),  A.A. and X.W. designed the experiment(s), A.A. and X.W. analyzed the results. Both authors written and reviewed the manuscript. 

\section*{Conflict of Interest}
The authors declare no conflict of interest.

\section*{Code Availability}
Both the datasets and the code required for executing the algorithm have been included as part of the supplementary material. The algorithms are fully implemented using the Python programming language, and we have provided a detailed step-by-step guide for execution using Jupyter Notebook.

\section*{Tables}

\begin{table}[h]
    \centering
    \caption{Phase I Premise Testing: Summary of Nodes and Edges for Training Models from Different Sources: PubMed vs ChatGPT Datasets}
    \label{tab:model}    
    \begin{tabular}{lcccccccc}
        \hline
        \textbf{Dataset} & \multicolumn{2}{c}{\textbf{Alzheimer's}} & \multicolumn{2}{c}{\textbf{Cancer}} & \multicolumn{2}{c}{\textbf{Depression}} \\
        & \textbf{Nodes} & \textbf{Edges} & \textbf{Nodes} & \textbf{Edges} & \textbf{Nodes} & \textbf{Edges} \\
        \hline
        ChatGPT 2024      & 519 & 1194 & 559 & 1050 & 577 & 1108 \\
        PubMed 2010-2014 & 742 & 861 & 755 & 803 & 802 & 878 \\
        PubMed 2015-2019 & 774 & 940 & 828 & 977 & 775 & 826 \\
        PubMed 2020-2024 & 817 & 958 & 817 & 1030 & 790 & 809 \\
        \hline
    \end{tabular}
\end{table}

\begin{table}[ht]
    \centering
    \caption{Phase II Premise Testing: Summary of Ratio Means for Different Diseases and Datasets}
    \label{tab:ratio_means}
    \begin{tabular}{lcccccccccccc}
        \toprule
        \textbf{} & \multicolumn{4}{c}{\textbf{Alzheimer's}} & \multicolumn{4}{c}{\textbf{Cancer}} & \multicolumn{4}{c}{\textbf{Depression}} \\
        \cmidrule(lr){2-5} \cmidrule(lr){6-9} \cmidrule(lr){10-13}
        & GPT & Pub 10-14 &  15-19 & 20-24 & GPT & 10-14 & 15-19 & 20-24 & GPT & 10-14 & 15-19 & 20-24 \\
        \midrule
        F-1 Means & 0.27 & 0.15 & 0.14 & 0.15 & 0.27 & 0.14 & 0.15 & 0.17 & 0.28 & 0.10 & 0.13 & 0.11 \\
        F-2 Means & 0.30 & 0.16 & 0.15 & 0.15 & 0.29 & 0.14 & 0.15 & 0.17 & 0.30 & 0.11 & 0.11 & 0.10 \\
        F-3 Means & 0.30 & 0.16 & 0.15 & 0.14 & 0.26 & 0.15 & 0.16 & 0.15 & 0.32 & 0.11 & 0.11 & 0.11 \\
        F-4 Means & 0.28 & 0.16 & 0.15 & 0.15 & 0.25 & 0.14 & 0.15 & 0.15 & 0.27 & 0.11 & 0.11 & 0.11 \\
        F-5 Means & 0.28 & 0.16 & 0.14 & 0.14 & 0.25 & 0.13 & 0.16 & 0.15 & 0.30 & 0.11 & 0.11 & 0.11 \\
        F-6 Means & 0.29 & 0.16 & 0.15 & 0.14 & 0.28 & 0.14 & 0.16 & 0.17 & 0.30 & 0.11 & 0.09 & 0.11 \\
        \bottomrule
    \end{tabular}
\end{table}

\begin{table}[ht]
    \centering
    \caption{Model Calibration: Summary of Lower and Upper Bound Ranges for Different Diseases and Datasets by Year Periods: 2010-2014, 2015-2019, 2020-2024}
    \label{tab:bound_ranges}
    \begin{tabular}{lcccccccccccc}
        \toprule
        \textbf{} & \multicolumn{4}{c}{\textbf{Alzheimer's}} & \multicolumn{4}{c}{\textbf{Cancer}} & \multicolumn{4}{c}{\textbf{Depression}} \\
        \cmidrule(lr){2-5} \cmidrule(lr){6-9} \cmidrule(lr){10-13}
        & GPT & Pub 10-14 & 15-19 & 20-24 & GPT & 10-14 & 15-19 & 20-24 & GPT & 10-14 & 15-19 & 20-24 \\
        \midrule
        Lower & 0.27 & 0.15 & 0.14 & 0.14 & 0.25 & 0.13 & 0.15 & 0.15 & 0.28 & 0.10 & 0.09 & 0.10 \\
        Upper & 0.30 & 0.16 & 0.15 & 0.15 & 0.29 & 0.15 & 0.16 & 0.17 & 0.32 & 0.11 & 0.13 & 0.11 \\
        \bottomrule
    \end{tabular}
\end{table}

\begin{table}[h]
    \centering
    \caption{Multi-Mode Experiments: xFakeSci Performance Evaluation for Recent Publications (2019 - 2024)}
    \label{tab:f1-by-disease} 
        \begin{tabular}{clrrrrr}
        \toprule
        Classifier Used & Disease Dataset Tested & TP & FP & FN & TN & F1 Score \\
        \bottomrule
        xFakeSci  & Depression & 50 & 25 & 0 & 25 & 80.00\% \\
        \bottomrule
        xFakeSci  & Cancer & 50 & 9 & 0 & 41 & 91.74\% \\
        \bottomrule
        xFakeSci  & Alzheimer's & 50 & 12 & 0 & 38 & 89.29\% \\
        \bottomrule
        \end{tabular}
\end{table}

\begin{table}[htbp]
\centering
\caption{Multi-Mode Experiments: F1 Classification Scores From Most Recent Publications }
\label{tab:f1_scores-most-recent}
    \begin{tabular}{cccc}
    \toprule
    Classifier / Publication (2020-2024) & Depression  & Cancer  & Alzheimer's \\
    \midrule
    xFakeSci & 80\% & 91\% & 89\% \\
    Naive Bayes & 52\% & 43\% & 47\% \\
    Linear SVM & 39\% & 51\% & 50\% \\
    Classical SVM & 44\% & 47\% & 41\% \\
    Logistic Regression & 44\% & 46\% & 49\% \\
    \bottomrule
    \end{tabular}
\end{table}

\begin{table}[htbp]
\centering
\caption{Multi-Mode Experiments: F1 Classification Scores By Older Periods}
\label{tab:f1_scores}
    \begin{tabular}{cccc}
    \toprule
    Publication Period (2014-2019) & Depression & Cancer & Alzheimer's \\ \hline
    xFakeSci & 80\% & 92\% & 90\% \\
    Naive Bayes & 50\% & 43\% & 44\% \\
    Linear SVM & 40\% & 51\% & 48\% \\
    Classical SVM & 44\% & 47\% & 40\% \\
    Logistic Regression & 45\% & 47\% & 46\% \\
    \midrule
    Publication Period (2010-2014) & Depression & Cancer & Alzheimer's \\ \hline
    xFakeSci & 80\% & 94\% & 89\% \\
    Naive Bayes & 51\% & 46\% & 45\% \\
    Linear SVM & 38\% & 52\% & 48\% \\
    Classical SVM & 42\% & 50\% & 41\% \\
    Logistic Regression & 42\% & 50\% & 48\% \\
    \bottomrule
    \end{tabular}
\end{table}


\clearpage

\section*{Figures}

\clearpage

\bibliography{references}

\begin{thebibliography}{10}
\urlstyle{rm}
\expandafter\ifx\csname url\endcsname\relax
  \def\url#1{\texttt{#1}}\fi
\expandafter\ifx\csname urlprefix\endcsname\relax\def\urlprefix{URL }\fi
\expandafter\ifx\csname doiprefix\endcsname\relax\def\doiprefix{DOI: }\fi
\providecommand{\bibinfo}[2]{#2}
\providecommand{\eprint}[2][]{\url{#2}}

\bibitem{chatgpt}
\bibinfo{title}{Chatgpt}.
\newblock \bibinfo{howpublished}{Online: \url{https://chat.openai.com}}
  (\bibinfo{year}{2023}).
\newblock \bibinfo{note}{Accessed August 15, 2023}.

\bibitem{synnestvedt2005citespace}
\bibinfo{author}{Synnestvedt, M.~B.}, \bibinfo{author}{Chen, C.} \&
  \bibinfo{author}{Holmes, J.~H.}
\newblock \bibinfo{title}{Citespace ii: visualization and knowledge discovery
  in bibliographic databases}.
\newblock In \emph{\bibinfo{booktitle}{AMIA annual symposium proceedings}},
  vol. \bibinfo{volume}{2005}, \bibinfo{pages}{724}
  (\bibinfo{organization}{American Medical Informatics Association},
  \bibinfo{year}{2005}).

\bibitem{holzinger2013graph}
\bibinfo{author}{Holzinger, A.} \emph{et~al.}
\newblock \bibinfo{title}{On graph entropy measures for knowledge discovery
  from publication network data}.
\newblock In \emph{\bibinfo{booktitle}{Availability, Reliability, and Security
  in Information Systems and HCI: IFIP WG 8.4, 8.9, TC 5 International
  Cross-Domain Conference, CD-ARES 2013, Regensburg, Germany, September 2-6,
  2013. Proceedings 8}}, \bibinfo{pages}{354--362}
  (\bibinfo{organization}{Springer}, \bibinfo{year}{2013}).

\bibitem{usai2018knowledge}
\bibinfo{author}{Usai, A.}, \bibinfo{author}{Pironti, M.},
  \bibinfo{author}{Mital, M.} \& \bibinfo{author}{Aouina~Mejri, C.}
\newblock \bibinfo{journal}{\bibinfo{title}{Knowledge discovery out of text
  data: a systematic review via text mining}}.
\newblock {\emph{\JournalTitle{Journal of knowledge management}}}
  \textbf{\bibinfo{volume}{22}}, \bibinfo{pages}{1471--1488}
  (\bibinfo{year}{2018}).

\bibitem{thaler2015fish}
\bibinfo{author}{Thaler, A.~D.} \& \bibinfo{author}{Shiffman, D.}
\newblock \bibinfo{journal}{\bibinfo{title}{Fish tales: Combating fake science
  in popular media}}.
\newblock {\emph{\JournalTitle{Ocean \& Coastal Management}}}
  \textbf{\bibinfo{volume}{115}}, \bibinfo{pages}{88--91}
  (\bibinfo{year}{2015}).

\bibitem{hopf2019fake}
\bibinfo{author}{Hopf, H.}, \bibinfo{author}{Krief, A.},
  \bibinfo{author}{Mehta, G.} \& \bibinfo{author}{Matlin, S.~A.}
\newblock \bibinfo{journal}{\bibinfo{title}{Fake science and the knowledge
  crisis: ignorance can be fatal}}.
\newblock {\emph{\JournalTitle{Royal Society open science}}}
  \textbf{\bibinfo{volume}{6}}, \bibinfo{pages}{190161} (\bibinfo{year}{2019}).

\bibitem{ho2022let}
\bibinfo{author}{Ho, S.~S.}, \bibinfo{author}{Goh, T.~J.} \&
  \bibinfo{author}{Leung, Y.~W.}
\newblock \bibinfo{journal}{\bibinfo{title}{Let’s nab fake science news:
  Predicting scientists’ support for interventions using the influence of
  presumed media influence model}}.
\newblock {\emph{\JournalTitle{Journalism}}} \textbf{\bibinfo{volume}{23}},
  \bibinfo{pages}{910--928} (\bibinfo{year}{2022}).

\bibitem{frederickson2022addressing}
\bibinfo{author}{Frederickson, R.~M.} \& \bibinfo{author}{Herzog, R.~W.}
\newblock \bibinfo{journal}{\bibinfo{title}{Addressing the big business of fake
  science}}.
\newblock {\emph{\JournalTitle{Molecular Therapy}}}
  \textbf{\bibinfo{volume}{30}}, \bibinfo{pages}{2390} (\bibinfo{year}{2022}).

\bibitem{rocha2021impact}
\bibinfo{author}{Rocha, Y.~M.} \emph{et~al.}
\newblock \bibinfo{journal}{\bibinfo{title}{The impact of fake news on social
  media and its influence on health during the covid-19 pandemic: A systematic
  review}}.
\newblock {\emph{\JournalTitle{Journal of Public Health}}}
  \bibinfo{pages}{1--10} (\bibinfo{year}{2021}).

\bibitem{walter2021evaluating}
\bibinfo{author}{Walter, N.}, \bibinfo{author}{Brooks, J.~J.},
  \bibinfo{author}{Saucier, C.~J.} \& \bibinfo{author}{Suresh, S.}
\newblock \bibinfo{journal}{\bibinfo{title}{Evaluating the impact of attempts
  to correct health misinformation on social media: A meta-analysis}}.
\newblock {\emph{\JournalTitle{Health communication}}}
  \textbf{\bibinfo{volume}{36}}, \bibinfo{pages}{1776--1784}
  (\bibinfo{year}{2021}).

\bibitem{loomba2021measuring}
\bibinfo{author}{Loomba, S.}, \bibinfo{author}{de~Figueiredo, A.},
  \bibinfo{author}{Piatek, S.~J.}, \bibinfo{author}{de~Graaf, K.} \&
  \bibinfo{author}{Larson, H.~J.}
\newblock \bibinfo{journal}{\bibinfo{title}{Measuring the impact of covid-19
  vaccine misinformation on vaccination intent in the uk and usa}}.
\newblock {\emph{\JournalTitle{Nature human behaviour}}}
  \textbf{\bibinfo{volume}{5}}, \bibinfo{pages}{337--348}
  (\bibinfo{year}{2021}).

\bibitem{lewandowsky2012misinformation}
\bibinfo{author}{Lewandowsky, S.}, \bibinfo{author}{Ecker, U.~K.},
  \bibinfo{author}{Seifert, C.~M.}, \bibinfo{author}{Schwarz, N.} \&
  \bibinfo{author}{Cook, J.}
\newblock \bibinfo{journal}{\bibinfo{title}{Misinformation and its correction:
  Continued influence and successful debiasing}}.
\newblock {\emph{\JournalTitle{Psychological science in the public interest}}}
  \textbf{\bibinfo{volume}{13}}, \bibinfo{pages}{106--131}
  (\bibinfo{year}{2012}).

\bibitem{myers2009misinformation}
\bibinfo{author}{Myers, M.} \& \bibinfo{author}{Pineda, D.}
\newblock \bibinfo{journal}{\bibinfo{title}{Misinformation about vaccines}}.
\newblock {\emph{\JournalTitle{Vaccines for biodefense and emerging and
  neglected diseases}}} \bibinfo{pages}{255--270} (\bibinfo{year}{2009}).

\bibitem{dailymail}
\bibinfo{author}{Matthews, S.} \& \bibinfo{author}{Spencer, B.}
\newblock \bibinfo{title}{Government orders review into vitamin d's role in
  covid-19}.
\newblock \bibinfo{howpublished}{Online:
  \url{https://www.dailymail.co.uk/news/article-8432321/Government-orders-review-vitamin-D-role-Covid-19.html}}
  (\bibinfo{year}{2020}).
\newblock \bibinfo{note}{Accessed on April 13, 2024}.

\bibitem{abdeen2021fighting}
\bibinfo{author}{Abdeen, M.~A.}, \bibinfo{author}{Hamed, A.~A.} \&
  \bibinfo{author}{Wu, X.}
\newblock \bibinfo{journal}{\bibinfo{title}{Fighting the covid-19 infodemic in
  news articles and false publications: The neonet text classifier, a
  supervised machine learning algorithm}}.
\newblock {\emph{\JournalTitle{Applied Sciences}}}
  \textbf{\bibinfo{volume}{11}}, \bibinfo{pages}{7265} (\bibinfo{year}{2021}).

\bibitem{hamed2024safeguarding}
\bibinfo{author}{Hamed, A.~A.}, \bibinfo{author}{Zachara-Szymanska, M.} \&
  \bibinfo{author}{Wu, X.}
\newblock \bibinfo{journal}{\bibinfo{title}{Safeguarding authenticity for
  mitigating the harms of generative ai: Issues, research agenda, and policies
  for detection, fact-checking, and ethical ai}}.
\newblock {\emph{\JournalTitle{iScience}}} \textbf{\bibinfo{volume}{27}},
  \bibinfo{pages}{108782},
  \doiprefix\url{https://doi.org/10.1016/j.isci.2024.108782}
  (\bibinfo{year}{2024}).

\bibitem{eysenbach2023role}
\bibinfo{author}{Eysenbach, G.} \emph{et~al.}
\newblock \bibinfo{journal}{\bibinfo{title}{The role of chatgpt, generative
  language models, and artificial intelligence in medical education: a
  conversation with chatgpt and a call for papers}}.
\newblock {\emph{\JournalTitle{JMIR Medical Education}}}
  \textbf{\bibinfo{volume}{9}}, \bibinfo{pages}{e46885} (\bibinfo{year}{2023}).

\bibitem{ieee_si_ed}
\bibinfo{title}{{IEEE} special issue on education in the world of {ChatGPT} and
  other generative {AI}}.
\newblock \bibinfo{howpublished}{Online:
  \url{https://ieee-edusociety.org/ieee-special-issue-education-world-chatgpt-and-other-generative-ai}}
  (\bibinfo{year}{2023}).
\newblock \bibinfo{note}{Accessed April 13, 2024}.

\bibitem{finc_innov}
\bibinfo{title}{Financial innovation}.
\newblock \bibinfo{howpublished}{Online:
  \url{https://jfin-swufe.springeropen.com/special-issue---chatgpt-and-generative-ai-in-finance}}
  (\bibinfo{year}{2023}).
\newblock \bibinfo{note}{Accessed April 13, 2024}.

\bibitem{mdpi_si_langs}
\bibinfo{title}{Special issue "language generation with pretrained models"}.
\newblock \bibinfo{howpublished}{Online:
  \url{https://www.mdpi.com/journal/languages/special_issues/K1Z08ODH6V}}
  (\bibinfo{year}{Year}).
\newblock \bibinfo{note}{Accessed April 13, 2023}.

\bibitem{oxford_sfi}
\bibinfo{title}{Call for papers for the special focus issue on {ChatGPT} and
  large language models ({LLMs}) in biomedicine and health}.
\newblock
  \bibinfo{howpublished}{\url{https://academic.oup.com/jamia/pages/call-for-papers-for-special-focus-issue}}
  (\bibinfo{year}{Year}).
\newblock \bibinfo{note}{Accessed July 4, 2023}.

\bibitem{jmir_glm}
\bibinfo{author}{Leung, T.~I.}, \bibinfo{author}{de~Azevedo~Cardoso, T.},
  \bibinfo{author}{Mavragani, A.} \& \bibinfo{author}{Eysenbach, G.}
\newblock \bibinfo{journal}{\bibinfo{title}{Best practices for using ai tools
  as an author, peer reviewer, or editor}}.
\newblock {\emph{\JournalTitle{J Med Internet Res}}}
  \textbf{\bibinfo{volume}{25}}, \bibinfo{pages}{e51584},
  \doiprefix\url{10.2196/51584} (\bibinfo{year}{2023}).

\bibitem{pnas_2023}
\bibinfo{author}{Null, N.}
\newblock \bibinfo{journal}{\bibinfo{title}{The {PNAS} journals outline their
  policies for {ChatGPT} and generative {AI}}}.
\newblock {\emph{\JournalTitle{PNAS Updates}}}
  \doiprefix\url{10.1073/pnas-updates.2023-02-21} (\bibinfo{year}{2023}).
\newblock \bibinfo{note}{Published online}.

\bibitem{brainard2023scientists}
\bibinfo{author}{Brainard, J.}
\newblock \bibinfo{journal}{\bibinfo{title}{As scientists explore ai-written
  text, journals hammer out policies}}.
\newblock {\emph{\JournalTitle{Science}}} \textbf{\bibinfo{volume}{379}},
  \bibinfo{pages}{740--741} (\bibinfo{year}{2023}).

\bibitem{fuster2023jacc}
\bibinfo{author}{Fuster, V.} \emph{et~al.}
\newblock \bibinfo{journal}{\bibinfo{title}{Jacc journals’ pathway forward
  with ai tools: The future is now}}.
\newblock {\emph{\JournalTitle{JACC: Advances}}} \textbf{\bibinfo{volume}{2}},
  \bibinfo{pages}{100296},
  \doiprefix\url{https://doi.org/10.1016/j.jacadv.2023.100296}
  (\bibinfo{year}{2023}).

\bibitem{flanagin2023nonhuman}
\bibinfo{author}{Flanagin, A.}, \bibinfo{author}{Bibbins-Domingo, K.},
  \bibinfo{author}{Berkwits, M.} \& \bibinfo{author}{Christiansen, S.~L.}
\newblock \bibinfo{journal}{\bibinfo{title}{Nonhuman ``authors'' and
  implications for the integrity of scientific publication and medical
  knowledge}}.
\newblock {\emph{\JournalTitle{Jama}}} \textbf{\bibinfo{volume}{329}},
  \bibinfo{pages}{637--639} (\bibinfo{year}{2023}).

\bibitem{chatgpt_plugins}
\bibinfo{title}{Chatgpt plugins}.
\newblock \bibinfo{howpublished}{Online: \url
  {https://openai.com/blog/chatgpt-plugins}} (\bibinfo{year}{2023}).
\newblock \bibinfo{note}{Accessed April 13, 2023}.

\bibitem{gilson2023does}
\bibinfo{author}{Gilson, A.} \emph{et~al.}
\newblock \bibinfo{journal}{\bibinfo{title}{How does chatgpt perform on the
  united states medical licensing examination? the implications of large
  language models for medical education and knowledge assessment}}.
\newblock {\emph{\JournalTitle{JMIR Medical Education}}}
  \textbf{\bibinfo{volume}{9}}, \bibinfo{pages}{e45312} (\bibinfo{year}{2023}).

\bibitem{chaka2023detecting}
\bibinfo{author}{Chaka, C.}
\newblock \bibinfo{journal}{\bibinfo{title}{Detecting ai content in responses
  generated by chatgpt, youchat, and chatsonic: The case of five ai content
  detection tools}}.
\newblock {\emph{\JournalTitle{Journal of Applied Learning and Teaching}}}
  \textbf{\bibinfo{volume}{6}} (\bibinfo{year}{2023}).

\bibitem{vapnik1999overview}
\bibinfo{author}{Vapnik, V.~N.}
\newblock \bibinfo{journal}{\bibinfo{title}{An overview of statistical learning
  theory}}.
\newblock {\emph{\JournalTitle{IEEE transactions on neural networks}}}
  \textbf{\bibinfo{volume}{10}}, \bibinfo{pages}{988--999}
  (\bibinfo{year}{1999}).

\bibitem{cing2023detecting}
\bibinfo{author}{Cingillioglu, I.}
\newblock \bibinfo{journal}{\bibinfo{title}{Detecting ai-generated essays: the
  chatgpt challenge}}.
\newblock {\emph{\JournalTitle{The International Journal of Information and
  Learning Technology}}} \textbf{\bibinfo{volume}{40}},
  \bibinfo{pages}{259--268} (\bibinfo{year}{2023}).

\bibitem{copyleaks}
\bibinfo{title}{Copyleaks: {AI} \& machine learning powered plagiarism
  checker}.
\newblock \bibinfo{howpublished}{Online: \url{https://copyleaks.com/}}.
\newblock \bibinfo{note}{Accessed April 13, 2024}.

\bibitem{crossplag}
\bibinfo{title}{Crossplag: Online plagiarism checker}.
\newblock \bibinfo{howpublished}{Online: \url{https://crossplag.com/}}.
\newblock \bibinfo{note}{Accessed April 13, 2024}.

\bibitem{elkhatat2023evaluating}
\bibinfo{author}{Elkhatat, A.~M.}, \bibinfo{author}{Elsaid, K.} \&
  \bibinfo{author}{Almeer, S.}
\newblock \bibinfo{journal}{\bibinfo{title}{Evaluating the efficacy of ai
  content detection tools in differentiating between human and ai-generated
  text}}.
\newblock {\emph{\JournalTitle{International Journal for Educational
  Integrity}}} \textbf{\bibinfo{volume}{19}}, \bibinfo{pages}{17}
  (\bibinfo{year}{2023}).

\bibitem{Andersone001568}
\bibinfo{author}{Anderson, N.} \emph{et~al.}
\newblock \bibinfo{journal}{\bibinfo{title}{Ai did not write this manuscript,
  or did it? can we trick the ai text detector into generated texts? the
  potential future of chatgpt and ai in sports \& exercise medicine manuscript
  generation}}.
\newblock {\emph{\JournalTitle{BMJ Open Sport \& Exercise Medicine}}}
  \textbf{\bibinfo{volume}{9}}, \doiprefix\url{10.1136/bmjsem-2023-001568}
  (\bibinfo{year}{2023}).

\bibitem{rashidi2023chatgpt}
\bibinfo{author}{Rashidi, H.~H.}, \bibinfo{author}{Fennell, B.~D.},
  \bibinfo{author}{Albahra, S.}, \bibinfo{author}{Hu, B.} \&
  \bibinfo{author}{Gorbett, T.}
\newblock \bibinfo{journal}{\bibinfo{title}{The chatgpt conundrum:
  Human-generated scientific manuscripts misidentified as ai creations by ai
  text detection tool}}.
\newblock {\emph{\JournalTitle{Journal of Pathology Informatics}}}
  \textbf{\bibinfo{volume}{14}}, \bibinfo{pages}{100342}
  (\bibinfo{year}{2023}).

\bibitem{pubmed-portal}
\bibinfo{author}{NLM, N. L. o.~M.}
\newblock \bibinfo{title}{National center of biotechnology information}.
\newblock \bibinfo{howpublished}{Online:
  \url{https://pubmed.ncbi.nlm.nih.gov/}}.
\newblock \bibinfo{note}{Accessed on January 25, 2024}.

\bibitem{wu2008top}
\bibinfo{author}{Wu, X.} \emph{et~al.}
\newblock \bibinfo{journal}{\bibinfo{title}{Top 10 algorithms in data mining}}.
\newblock {\emph{\JournalTitle{Knowledge and Information Systems}}}
  \textbf{\bibinfo{volume}{14}}, \bibinfo{pages}{1--37} (\bibinfo{year}{2008}).

\bibitem{scikit-learn}
\bibinfo{author}{Pedregosa, F.} \emph{et~al.}
\newblock \bibinfo{journal}{\bibinfo{title}{Scikit-learn: Machine learning in
  {P}ython}}.
\newblock {\emph{\JournalTitle{Journal of Machine Learning Research}}}
  \textbf{\bibinfo{volume}{12}}, \bibinfo{pages}{2825--2830}
  (\bibinfo{year}{2011}).

\bibitem{aizawa2003information}
\bibinfo{author}{Aizawa, A.}
\newblock \bibinfo{journal}{\bibinfo{title}{An information-theoretic
  perspective of tf--idf measures}}.
\newblock {\emph{\JournalTitle{Information Processing \& Management}}}
  \textbf{\bibinfo{volume}{39}}, \bibinfo{pages}{45--65}
  (\bibinfo{year}{2003}).

\bibitem{qaiser2018text}
\bibinfo{author}{Qaiser, S.} \& \bibinfo{author}{Ali, R.}
\newblock \bibinfo{journal}{\bibinfo{title}{Text mining: use of tf-idf to
  examine the relevance of words to documents}}.
\newblock {\emph{\JournalTitle{International Journal of Computer
  Applications}}} \textbf{\bibinfo{volume}{181}}, \bibinfo{pages}{25--29}
  (\bibinfo{year}{2018}).

\bibitem{ramos2003using}
\bibinfo{author}{Ramos, J.} \emph{et~al.}
\newblock \bibinfo{title}{Using tf-idf to determine word relevance in document
  queries}.
\newblock In \emph{\bibinfo{booktitle}{Proceedings of the first instructional
  conference on machine learning}}, vol. \bibinfo{volume}{242,{1}},
  \bibinfo{pages}{29--48} (\bibinfo{organization}{Citeseer},
  \bibinfo{year}{2003}).

\bibitem{trstenjak2014knn}
\bibinfo{author}{Trstenjak, B.}, \bibinfo{author}{Mikac, S.} \&
  \bibinfo{author}{Donko, D.}
\newblock \bibinfo{journal}{\bibinfo{title}{Knn with tf-idf based framework for
  text categorization}}.
\newblock {\emph{\JournalTitle{Procedia Engineering}}}
  \textbf{\bibinfo{volume}{69}}, \bibinfo{pages}{1356--1364}
  (\bibinfo{year}{2014}).

\bibitem{wu2008interpreting}
\bibinfo{author}{Wu, H.~C.}, \bibinfo{author}{Luk, R. W.~P.},
  \bibinfo{author}{Wong, K.~F.} \& \bibinfo{author}{Kwok, K.~L.}
\newblock \bibinfo{journal}{\bibinfo{title}{Interpreting tf-idf term weights as
  making relevance decisions}}.
\newblock {\emph{\JournalTitle{ACM Transactions on Information Systems
  (TOIS)}}} \textbf{\bibinfo{volume}{26}}, \bibinfo{pages}{1--37}
  (\bibinfo{year}{2008}).

\bibitem{zhang2011comparative}
\bibinfo{author}{Zhang, W.}, \bibinfo{author}{Yoshida, T.} \&
  \bibinfo{author}{Tang, X.}
\newblock \bibinfo{journal}{\bibinfo{title}{A comparative study of tf* idf, lsi
  and multi-words for text classification}}.
\newblock {\emph{\JournalTitle{Expert systems with applications}}}
  \textbf{\bibinfo{volume}{38}}, \bibinfo{pages}{2758--2765}
  (\bibinfo{year}{2011}).

\bibitem{tan2002use}
\bibinfo{author}{Tan, C.-M.}, \bibinfo{author}{Wang, Y.-F.} \&
  \bibinfo{author}{Lee, C.-D.}
\newblock \bibinfo{journal}{\bibinfo{title}{The use of bigrams to enhance text
  categorization}}.
\newblock {\emph{\JournalTitle{Information processing \& management}}}
  \textbf{\bibinfo{volume}{38}}, \bibinfo{pages}{529--546}
  (\bibinfo{year}{2002}).

\bibitem{hirst2007bigrams}
\bibinfo{author}{Hirst, G.} \& \bibinfo{author}{Feiguina, O.}
\newblock \bibinfo{journal}{\bibinfo{title}{Bigrams of syntactic labels for
  authorship discrimination of short texts}}.
\newblock {\emph{\JournalTitle{Literary and Linguistic Computing}}}
  \textbf{\bibinfo{volume}{22}}, \bibinfo{pages}{405--417}
  (\bibinfo{year}{2007}).

\bibitem{dorogovtsev2001giant}
\bibinfo{author}{Dorogovtsev, S.~N.}, \bibinfo{author}{Mendes, J. F.~F.} \&
  \bibinfo{author}{Samukhin, A.~N.}
\newblock \bibinfo{journal}{\bibinfo{title}{Giant strongly connected component
  of directed networks}}.
\newblock {\emph{\JournalTitle{Physical Review E}}}
  \textbf{\bibinfo{volume}{64}}, \bibinfo{pages}{025101}
  (\bibinfo{year}{2001}).

\bibitem{kitsak2018stability}
\bibinfo{author}{Kitsak, M.} \emph{et~al.}
\newblock \bibinfo{journal}{\bibinfo{title}{Stability of a giant connected
  component in a complex network}}.
\newblock {\emph{\JournalTitle{Physical Review E}}}
  \textbf{\bibinfo{volume}{97}}, \bibinfo{pages}{012309}
  (\bibinfo{year}{2018}).

\bibitem{Beygelzimer2005}
\bibinfo{author}{Beygelzimer, A.}, \bibinfo{author}{Grinstein, G.},
  \bibinfo{author}{Linsker, R.} \& \bibinfo{author}{Rish, I.}
\newblock \bibinfo{journal}{\bibinfo{title}{Improving network robustness by
  edge modification}}.
\newblock {\emph{\JournalTitle{Physica A: Statistical Mechanics and its
  Applications}}} \textbf{\bibinfo{volume}{357}},
  \doiprefix\url{10.1016/j.physa.2005.03.040} (\bibinfo{year}{2005}).

\bibitem{Zhang2017}
\bibinfo{author}{Zhang, G.}, \bibinfo{author}{Duan, H.} \&
  \bibinfo{author}{Zhou, J.}
\newblock \bibinfo{journal}{\bibinfo{title}{Network stability, connectivity and
  innovation output}}.
\newblock {\emph{\JournalTitle{Technological Forecasting and Social Change}}}
  \textbf{\bibinfo{volume}{114}},
  \doiprefix\url{10.1016/j.techfore.2016.09.004} (\bibinfo{year}{2017}).

\bibitem{Bellingeri2020}
\bibinfo{author}{Bellingeri, M.} \emph{et~al.}
\newblock \bibinfo{journal}{\bibinfo{title}{Link and node removal in real
  social networks: A review}}.
\newblock {\emph{\JournalTitle{Frontiers in Physics}}}
  \textbf{\bibinfo{volume}{8}}, \doiprefix\url{10.3389/fphy.2020.00228}
  (\bibinfo{year}{2020}).

\bibitem{genkin2007large}
\bibinfo{author}{Genkin, A.}, \bibinfo{author}{Lewis, D.~D.} \&
  \bibinfo{author}{Madigan, D.}
\newblock \bibinfo{journal}{\bibinfo{title}{Large-scale bayesian logistic
  regression for text categorization}}.
\newblock {\emph{\JournalTitle{technometrics}}} \textbf{\bibinfo{volume}{49}},
  \bibinfo{pages}{291--304} (\bibinfo{year}{2007}).

\bibitem{feng2017overfitting}
\bibinfo{author}{Feng, X.} \emph{et~al.}
\newblock \bibinfo{journal}{\bibinfo{title}{Overfitting reduction of text
  classification based on adabelm}}.
\newblock {\emph{\JournalTitle{Entropy}}} \textbf{\bibinfo{volume}{19}},
  \bibinfo{pages}{330} (\bibinfo{year}{2017}).

\bibitem{deng2019}
\bibinfo{author}{Deng, X.}, \bibinfo{author}{Li, Y.}, \bibinfo{author}{Weng,
  J.} \& \bibinfo{author}{Zhang, J.}
\newblock \bibinfo{journal}{\bibinfo{title}{Feature selection for text
  classification: A review}}.
\newblock {\emph{\JournalTitle{Multimedia Tools and Applications}}}
  \textbf{\bibinfo{volume}{78}}, \bibinfo{pages}{3797--3816},
  \doiprefix\url{10.1007/s11042-018-6083-5} (\bibinfo{year}{2019}).

\bibitem{khurana9691845_2023}
\bibinfo{author}{Khurana, A.} \& \bibinfo{author}{Verma, O.~P.}
\newblock \bibinfo{journal}{\bibinfo{title}{Optimal feature selection for
  imbalanced text classification}}.
\newblock {\emph{\JournalTitle{IEEE Transactions on Artificial Intelligence}}}
  \textbf{\bibinfo{volume}{4}}, \bibinfo{pages}{135--147},
  \doiprefix\url{10.1109/TAI.2022.3144651} (\bibinfo{year}{2023}).

\bibitem{conroy2023chatgpt}
\bibinfo{author}{Conroy, G.}
\newblock \bibinfo{journal}{\bibinfo{title}{How chatgpt and other ai tools
  could disrupt scientific publishing}}.
\newblock {\emph{\JournalTitle{Nature}}} \textbf{\bibinfo{volume}{622}},
  \bibinfo{pages}{234--236} (\bibinfo{year}{2023}).

\end{thebibliography}

\end{document}